\documentclass[lettersize,journal]{IEEEtran}
\usepackage{amsmath,amsfonts}
\usepackage[ruled,linesnumbered]{algorithm2e}
\usepackage{algorithmic}
\usepackage{array}
\usepackage[caption=false,font=normalsize,labelfont=sf,textfont=sf]{subfig}
\usepackage{textcomp}
\usepackage{stfloats}
\usepackage{url}
\usepackage{verbatim}
\usepackage{graphicx}

\usepackage{booktabs}
\usepackage{amssymb}
\usepackage{multirow}
\usepackage[normalem]{ulem}
\useunder{\uline}{\ul}{}
\usepackage{hyperref}
\usepackage{floatrow}
\newfloatcommand{capbtabbox}{table}[][\FBwidth]

\begin{document}

\title{End-to-End Streaming Video Temporal Action Segmentation with Reinforcement Learning}

\author{Jinrong Zhang, Wujun Wen, Shenglan Liu, Gao Huang, Yunheng Li, Qifeng Li, Lin Feng
        % <-this % stops a space
\thanks{Jinrong Zhang and Wujun Wen are equal contribution.}
\thanks{Jinrong Zhang is with the School of Control Science and Engineering, Dalian University of Technology, Dalian 116024, China (e-mail: zjr15272565639@mail.dlut.edu.cn).}
\thanks{Wujun Wen, Yunheng Li, Qifeng Li are with the School of Computer Science and Technology, Dalian University of Technology, Dalian 116024, China (e-mail: wujunwen@mail.dlut.edu.cn; liyunheng@mail.dlut.edu.cn; qifengli@mail.dlut.edu.cn).}
\thanks{Shenglan Liu is with the School of Innovation and Entrepreneurship, Dalian University of Technology, Dalian 116024, China (e-mail: liusl@dlut.edu.cn).}
\thanks{Gao Huang is with the Department of Automation, Tsinghua University, Beijing 100084, China (e-mail: gaohuang@tsinghua.edu.cn).}
\thanks{Lin Feng is with the School of Information and Communication Engineering, Dalian Minzu University, 116600 Dalian 116600 Liaoning, China, and also with the School of Innovation and Entrepreneurship, Dalian University of Technology, Dalian 116024, China (e-mail: fenglin@dlut.edu.cn).\\
(Corresponding author: Sheng-Lan Liu.)
}}

\markboth{Journal of \LaTeX\ Class Files,~Vol.~14, No.~8, August~2021}%
{Shell \MakeLowercase{\textit{et al.}}: A Sample Article Using IEEEtran.cls for IEEE Journals}


\maketitle

\begin{abstract}
The streaming temporal action segmentation (STAS) task, a supplementary task of temporal action segmentation (TAS), has not received adequate attention in the field of video understanding. Existing TAS methods are constrained to offline scenarios due to their heavy reliance on multimodal features and complete contextual information. The STAS task requires the model to classify each frame of the entire untrimmed video sequence clip by clip in time, thereby extending the applicability of TAS methods to online scenarios. However, directly applying existing TAS methods to SATS tasks results in significantly poor segmentation outcomes. In this paper, we thoroughly analyze the fundamental differences between STAS tasks and TAS tasks, attributing the severe performance degradation when transferring models to model bias and optimization dilemmas. We introduce an end-to-end streaming video temporal action segmentation model with reinforcement learning (SVTAS-RL). The end-to-end modeling method mitigates the modeling bias introduced by the change in task nature and enhances the feasibility of online solutions. Reinforcement learning is utilized to alleviate the optimization dilemma. Through extensive experiments, the SVTAS-RL model significantly outperforms existing STAS models and achieves competitive performance to the state-of-the-art TAS model on multiple datasets under the same evaluation criteria, demonstrating notable advantages on the ultra-long video dataset EGTEA. Code is available at https://github.com/Thinksky5124/SVTAS.
\end{abstract}


\section{Introduction}
\IEEEPARstart{S}{treaming} Temporal Action Segmentation (STAS), as a task with broad application prospects, has not yet received as much attention as Temporal Action Segmentation (TAS). Current TAS methods are confined to offline settings because they rely on multimodal features from complete videos, which involve multi-stage training and complex pipeline processing. Unlike TAS, which directly assigns labels to every frame of a complete video \cite{ding2022temporal}, STAS divides a full video into multiple continuous video clips and inputs them into the model in a streaming manner. In STAS, the model processes only one video clip at a time to compute its temporal segmentation result. Subsequently, it concatenates the segmentation results of all video clips to generate the final segmentation outcome for the complete video. The input method of streaming video clips brings broader application prospects to STAS. With the introduction of streaming video inputs and the reduction in duration of streaming video clips, online scenarios such as online teaching and live broadcasting become feasible. STAS not only offers a novel online solution for the field of temporal segmentation but also poses greater challenges.

\begin{figure*}[h]
  \centering
  \includegraphics[width=0.95\linewidth]{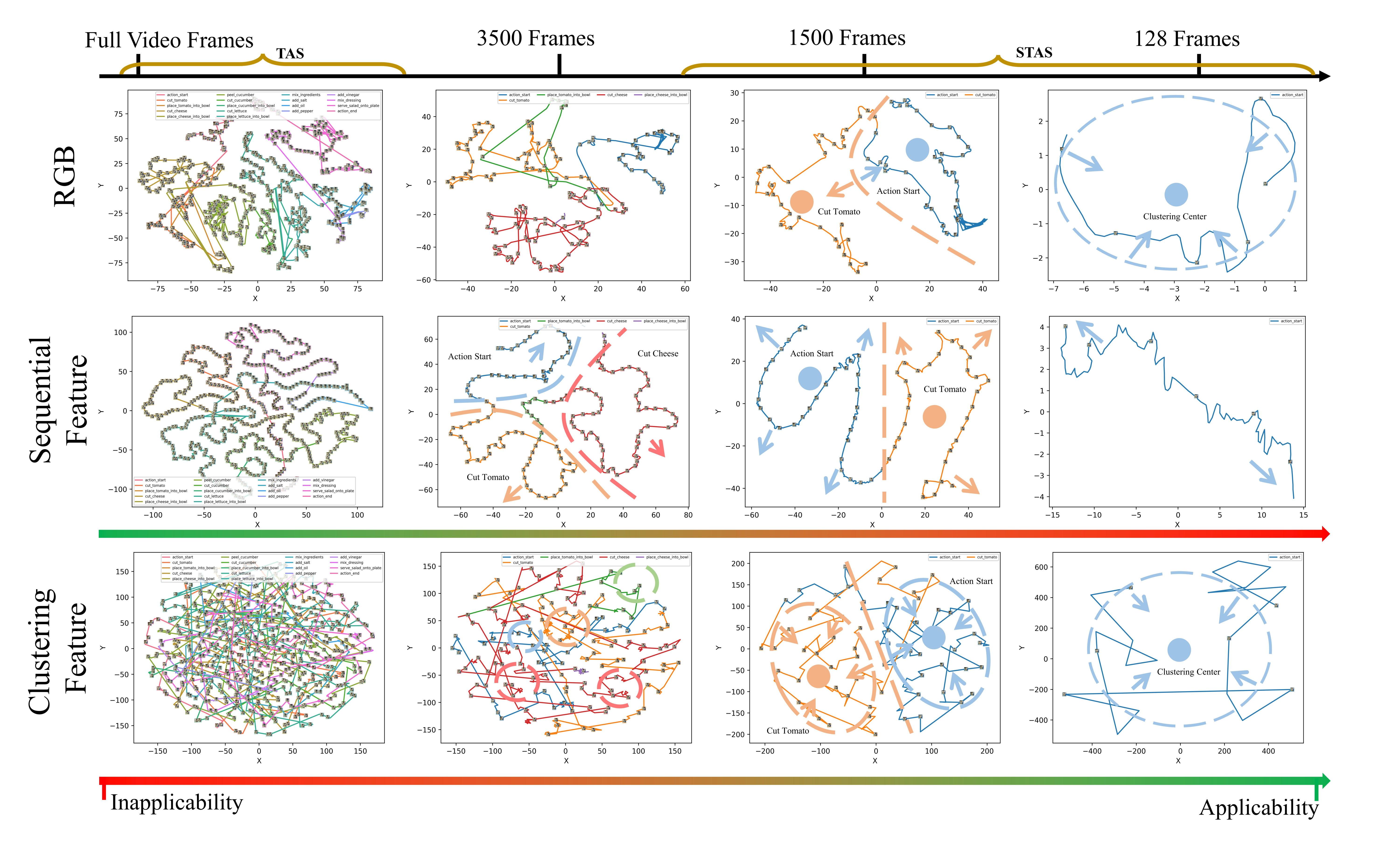}
  \caption{The phenomenon of modeling bias in TAS models migrating to STAS task. All visualization manifolds of the data are obtained through t-SNE. In the image, the horizontal axis represents the length of the video clip after cropping, with the complete video on the far right and progressively shorter lengths towards the left.The first row represents the manifolds of the RGB modality of video images, the second row depicts the manifolds of the TAS model sequential features, and the third row illustrates the manifolds of the clustering model features. As the duration of segmented video clips decreases, which means from the TAS task to the STAS task, we can observe: (a) the data manifold of the original video gradually transitions from a distorted line to a clustering swiss roll; (b) The shorter the segmented video clip, the less applicable the sequential model becomes, but clustering model performs the opposite.}
  \label{model_bias}
\end{figure*}

Although it appears that the only difference between STAS and TAS lies in the length of video processed in a single forward operation, transferring existing TAS methods to the STAS task results in significant performance degradation. This phenomenon prompts us to consider the fundamental differences between TAS and STAS. Current research treats TAS models as sequence-to-sequence transformation systems with designed feature extraction processes \cite{ding2022temporal}, termed the sequence paradigm. To study the impact of streaming video data input, we reduce and visualize both the original RGB data of complete videos and different lengths of streaming video clips, alongside the features extracted using the sequence paradigm, as shown in Fig. \ref{model_bias}. Complete RGB videos in high-dimensional space appear as a continuous twisted curve along the time dimension. As video clip length decreases, the sequential characteristics of the original video clips diminish. The features of cropped original video clips, resembling those extracted using a clustering paradigm, suggest better suitability for description by clustering paradigms (detailed in Section \ref{paradigm}). The sequential characteristics of sequential features do not change with the reduction in video clip length, which increasingly mismatches the data manifold of the original RGB data. This indicates a significant modeling bias between existing TAS methods and the STAS task. Additionally, the optimization objective for STAS is to maximize the integrity of action segments throughout the video sequence, while the optimization for existing TAS on individual streaming video clips focuses on minimizing frame-level classification loss within each clip. This leads to discrepancies between the gradients calculated by existing supervised optimization methods and the target gradients (See Section \ref{gradient_estimate}), a situation we refer to as an optimization dilemma problem. In summary, STAS is a more challenging task that existing TAS methods are ill-equipped to address. The challenges brought by STAS include: (a) the modeling bias severely limits the ability of the model to achieve sequence-to-sequence transformation; (b) the optimization dilemma problem of the training process for STAS models makes the model often falling into the local optimum; (c) streaming data inherently lacks future context information \cite{yudual}, which has an impact on the integrity of the action segment.

To tackle these challenges, we propose Streaming Video Temporal Action Segmentation with Reinforcement Learning (SVTAS-RL). Specifically, consider the video as an infinite video stream and perform action sequence segmentation directly on a limited length of time step, i.e., a clip of the video each time. At the end, all segmentation results are concatenated. Different from the TAS models, which generates sequential features by sliding window with step size of 1, SVTAS-RL directly extracts clustering features and segments action on current video clips. Our method eliminates modeling bias by aligning the modeling method with the raw data manifold. Similar to offline Automatic Speech Recognition (ASR) \cite{wang2019overview}, the optimum of each module does not necessarily mean global optimum \cite{zhang2019towards,graves2014towards} in training STAS. Our method can be trained end-to-end to get rid of tedious training process through limited length of time step, allowing global optimization and mitigating propagation of error. Additionally, it is not feasible to use full-sequence-based approaches such as post-processing \cite{li2021efficient} or multi-stage methods \cite{ahn2021refining} to avoid optimization dilemma. Moreover, current supervised optimization strategies in TAS task are unable to calculate the corresponding gradient of the optimization objective. Inspired by Reinforcement Learning from Human Feedback (RLHF), Reinforcement Learning (RL) can be used for online training and estimating the gradient of the optimization objective via cumulative expectation to overcome optimization dilemma \cite{williams1992simple,sutton1999policy}, which is very suitable for STAS learning. We regard STAS as a sequential decision-making task based on clustering and propose two distinct RL learning strategies to estimate the gradient: Monte Carlo Episodic REINFORCE Learning and Temporal Difference Actor-Critic Learning.

In summary, this paper presents three main contributions:
\begin{itemize}
  \item We display the phenomenon of modeling bias for the first time that occurs when TAS models migrate to STAS task. And, we propose SVTAS-RL model that aligns the modeling method with the raw data manifold to eliminate modeling bias.
  \item We are the first to combine RL with STAS to alleviate optimization dilemma by estimating gradient corresponding to the integrity of action of the full sequence. And, we propose two RL learning algorithms suitable for STAS.
  \item Extensive experiments show that SVTAS-RL which our proposed has achieved competitive performance to the State-Of-The-Art (SOTA) model of TAS on multiple datasets under the same evaluation. Moreover, our approach completely outperforms the existing STAS model and shows great performance improvement on the ultra-long video dataset EGTEA.
\end{itemize}

\section{Related Work}
\subsection{TAS Model based on Sequence-to-sequence Transformation}
Recently TAS and STAS models are belonged to this category. LBS \cite{li2021efficient} that is post-processing method to improve model performance; and HASR \cite{ahn2021refining} that using multi-stage segmentation to improve the segmentation performance of the full video sequence, and so on \cite{lea2017temporal,lea2016segmental,lea2017temporal,lei2018temporal,farha2019ms,yi2021asformer,li2020ms,singhania2021coarse,dong2022double, li2023involving}. These methods are based on the assumption that information for the full video sequence can be obtained, which can only apply offline scenario. Recently, research on improving training methods to alleviate optimization dilemma has mainly focused on adding auxiliary loss functions in TAS supervised training \cite{asrf,BCN}, such as T-MSE \cite{farha2019ms}, which uses a smooth loss to alleviate over-segmentation issues. However, this is an indirect loss function that optimizes the STAS optimization objective, and excessive smoothing greatly reduces model performance. Furthermore, as a phenomenon that has not been discovered by scholars, modeling bias also seriously affects the performance of TAS models on STAS task.

\subsection{Online Video Understanding}
To the best of our knowledge, the online video understanding tasks related to STAS are Action Recognition (AR), Online Temporal Action Localization (OTAL) and Online Action Detection (OAD). However, like the difference between semantic segmentation and object detection, the purpose of OTAL and OAD aims to detect action instances \cite{huang2019spatial}, TAS and STAS aim to achieve frame-level classification. STAS-related methods are not migratability and comparability (See Tab.\ref{compare_task}). The current OAS method \cite{ghoddoosian2022weakly,kumar2022unsupervised} has significant performance gap from our proposed SVTAS-RL. ETSN \cite{kang2022efficient} is the first online TAS method, which proposes a dual-stream action segmentation pipeline that can effectively learn motion and spatial information and perform online TAS. However, ETSN also has significant performance gap from current TAS models.

\subsection{RL in Video Analysis}
RL \cite{bellman1954theory} can tackle sequential decision-making in dynamic programming. DSN \cite{zhou2018deep} is a framework of reinforcement learning to resolve video summary task, which regards Video Summary (VS) task as sequential frames selection from the video by the agent, and uses the REINFOECE algorithm for training; In Temporal Action Detection (TAD) task, recent studies \cite{yeung2016end,wang2019language} that use RL consider detecting an action as a sequential searching decision in hole video by agent; and so on \cite{zhang2021real}. In STAS task, segmenting sequential video clips is also a sequential decision-making process. RL has the capability to optimize the overall decision sequence, analogous to optimizing the entire sequence of a video.

\begin{figure*}[t]
  \centering
  \includegraphics[width=\linewidth]{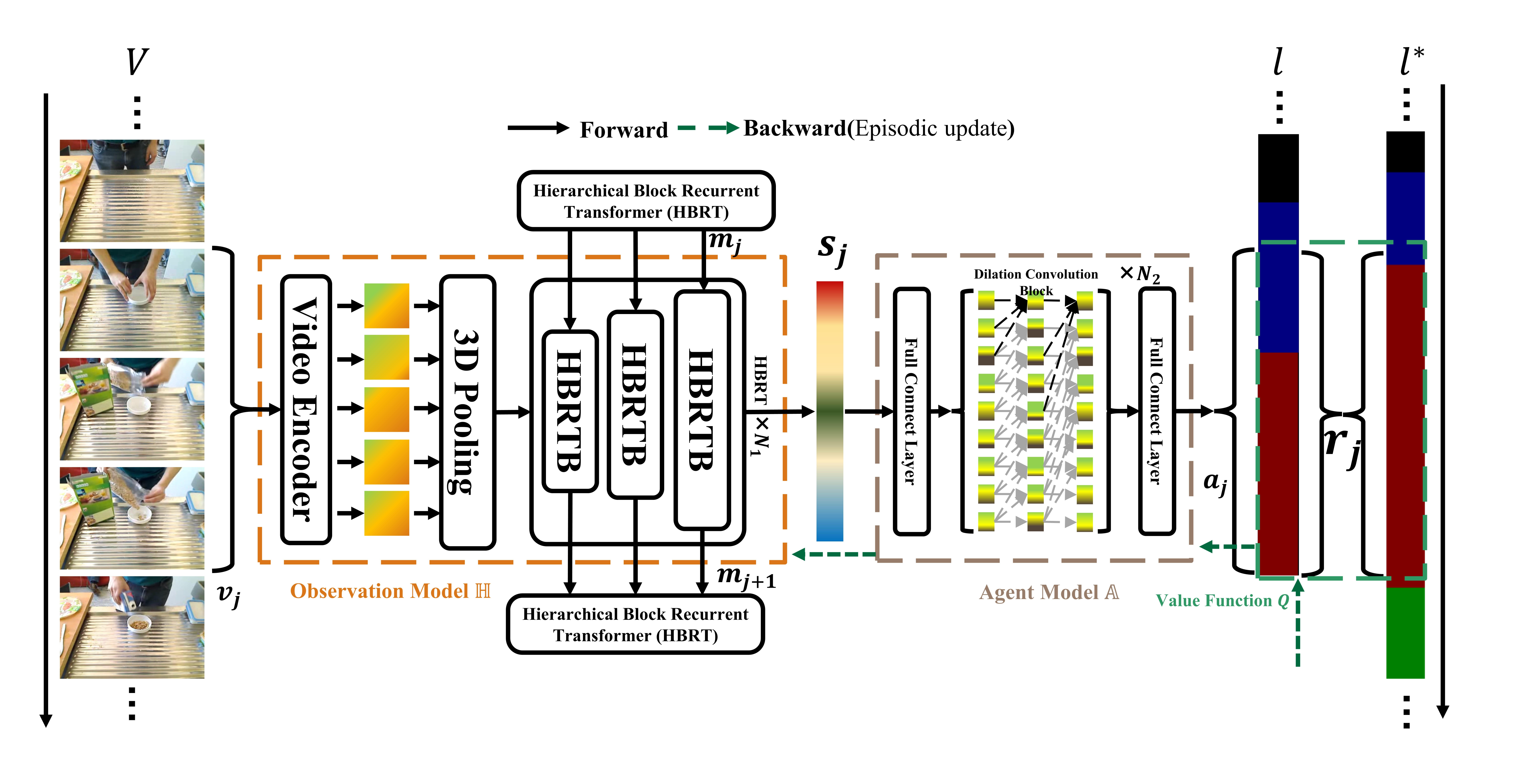}
  \caption{Overview of SVTAS-RL model. To train the SVTAS-RL model, we first sample a video clip $v_j$, which is then parsed by the observation model $\mathbb{H}$ to yield the current state $s_j$. Subsequently, the agent $\mathbb{A}$ makes a decision $a_j$ based on the current state $s_j$, and the decision is evaluated by $Q$ to obtain a reward $r_j$.}
  \label{model_fig}
\end{figure*}

\section{Method}
The inference process for a SVTAS-RL model can be regarded as a sequential decision-making system based on clustering that emulates a robotic agent observing current environment (video clip) as a state, then traversing a sequence of states (clustering action segment feature) and making decisions (segment action) simultaneously. The quality of the decision is evaluated through feedback on the integrity of action segment for the full video.

\subsection{Task Definition}
We regard a video $ V = \{ x_i | i=0, \cdots, T-1 \} $ as an ordered collection of image frames $x_i$, where $T$ denotes the total number of frames in the video. Each frame is assigned a corresponding label, denoted as $l_i^* \in \{ 0, \cdots, C-1\} $, where $C$ represents the total number of action categories. The model's predicted result is denoted by $l_i \in \{ 0, \cdots, C-1 \} $. For STAS task of feature, We perform non-overlapping stream sampling of the feature. For STAS task, We perform non-overlapping stream sampling of the video frame to ensure its efficient processing. After sampling, a feature clip or video clip $v_j$ will be fed into model to yield segment result $[l_{j \cdot L}, \cdots, l_{j \cdot L + L}]$, where $j=0, \cdots, \lceil \frac{T}{L} \rceil$ and $L$ is the length of clip. Repeat the process and collect the results in each iteration until the end of the video. To model the task as a sequential decision-making problem, we define features of video clip as a state $s_j$. Accordingly, we define segment result of $v_j$ as an action $a_j = [l_{j \cdot L}, \cdots, l_{j \cdot L + L}]$ and the action space is $a_j \in \mathbb{R}^{C \times L}$. Specifically, the agent model is defined as $\mathbb{A}(s_j; \theta)$, parameterized by $\theta$, the observation model as $\mathbb{H}(v_j, m_j; \phi)$, parameterized by $\phi$, and the value function as $Q(a^*_j, a_j)$, where $m_j$ is the historical information about the observation.

\subsection{Architecture}
When our proposed SVTAS-RL model is trained by supervised training methods instead of reinforcement learning method, we call it SVTAS.

\subsubsection{Observation Model}
As shown in Fig.\ref{model_fig}, the observation model $\mathbb{H}$, parameterized by $\phi$, observes a video clip at each time step. It encodes the video clip into a feature state $s_j$ and provides $s_j$ as input to the agent model. Importantly, given the high degree of video information redundancy, we leverage two common pre-processing techniques in RL and video understanding, namely frame stacking \cite{zhang2022improved_rl_frame_stacking} and frame skipping \cite{2013Playing_rl_frame_skipping}, to select an appropriate video clip $v_j$ length as a state $s_j$, which we formalize as $v_j \in \{ [x_j, x_{j+p}, \cdots, x_{j+L}] \}$, $L = k \times p$, where $k$ is the number of stacked frames, $p$ is the number of skipped frames, $L$ is the length of video clip. To extract rich information from video clip, we have employed the video swin transformer \cite{swin3d}, an action recognition model, as Video Encoder (VE) to observe the current video clip, and HBRT which inspired by Block-Recurrent Transformers \cite{block_recurrent_transformer} (BRT) to memorize history information and fuse the current video clip information. Notably, HBRT can be used as a model for STAS alone.

\begin{algorithm}[htbp]
  \SetKwData{Left}{left}\SetKwData{This}{this}\SetKwData{Up}{up}
  \SetKwFunction{Union}{Union}\SetKwFunction{FindCompress}{FindCompress}
  \SetKwInOut{Input}{input}\SetKwInOut{Output}{output}

  \Input{Image Sequence $I_j$, Video Encoder Model $VE(\cdot)$.}
  \Output{Clustering Feature $F^*_j$}
  \BlankLine
  $I_j = [x_{j * L}, \cdots, x_{j * L + k*p}]$\;
  $F_j = VE(I_j), F_j \in \mathbb{R}^{k * D * H * W}$, $H$ means image height, $W$ means image width, $D$ is the dimension of information\;
  $F^*_j = Pool3D_1(F_j), F^*_j \in \mathbb{R}^{k * D}$\;
  \caption{Clustering Paradigm Algorithm}
  \label{clustering_algorithm}
\end{algorithm}

\begin{algorithm}[htbp]
  \SetKwData{Left}{left}\SetKwData{This}{this}\SetKwData{Up}{up}
  \SetKwFunction{Union}{Union}\SetKwFunction{FindCompress}{FindCompress}
  \SetKwInOut{Input}{input}\SetKwInOut{Output}{output}

  \Input{Image Sequence $I_j$, Video Encoder Model $VE(\cdot)$.}
  \Output{Sequential Feature $F^*_j$}
  \BlankLine
  \For{$b\leftarrow 0$ \KwTo $k$}{
    $I_j = [x_{j * L + b * p - \lceil \frac{k * p}{2} \rceil}, \cdots, x_{j *L + + b * p + \lceil \frac{k * p}{2} \rceil}]$\;
    $F_j = VE(I_j), F_j \in \mathbb{R}^{k * D * H * W}$, $H$ means image height, $W$ means image width, $D$ is the dimension of information\;
    $f_{j * L + b * p} = Pool3D_2(F_j), f_{j * L} \in \mathbb{R}^{1 * D}$\;
  }
  $F^*_{j} = [f_{j * L}, \cdots, f_{j * L + k * p}], F^*_j \in \mathbb{R}^{k * D}$\;
  \caption{Sequential Paradigm Algorithm}
  \label{sequential_algorithm}
\end{algorithm}

\begin{figure}[h]
  \centering
  \includegraphics[width=1.\linewidth]{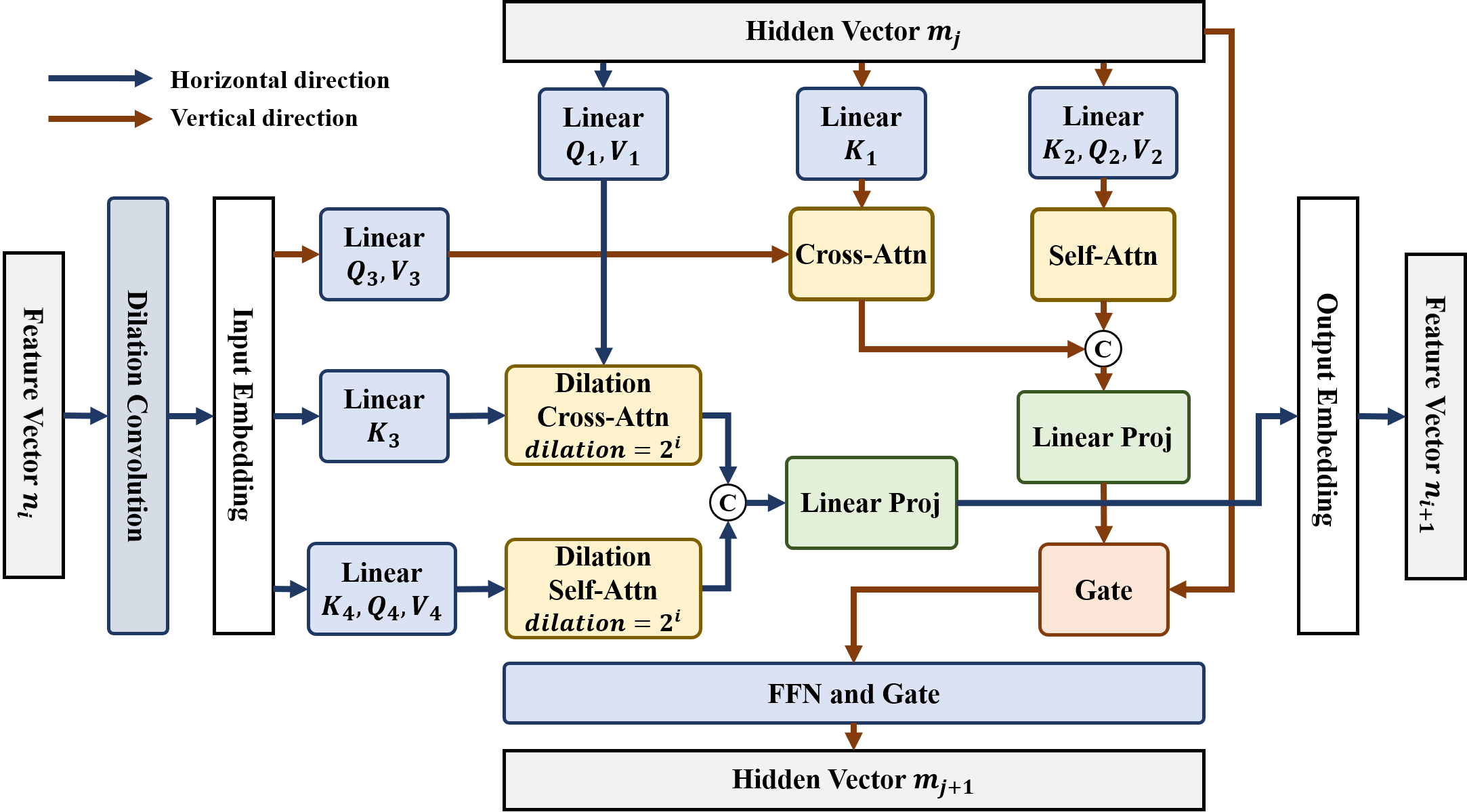}
  \caption{Overview of Hierarchical Block Recurrent Transformer (HBRT).}
  \label{hbrt_fig}
\end{figure}

\subsubsection{Clustering Paradigm}
\label{paradigm}
In order to eliminate modeling bias, we use clustering paradigm to build our model instead of sequential paradigm. We define the clustering paradigm and the sequential paradigm by mathematical notation for clear (See Algorithm.\ref{clustering_algorithm} and Algorithm.\ref{sequential_algorithm}).

\subsubsection{Hierarchical Block Recurrent Transformer}
The HBRT architecture, illustrated in Fig.\ref{hbrt_fig}, receives the output $f_j$ of VE and historical information $m_j \in \mathbb{R}^{D \times M}$ as input, and produces the state $s_j$ and updated historical information $m_{j+1} \in \mathbb{R}^{D \times M}$ as output, where $D$ is the dimension of information and $M$ is the length of information. Before feeding $f_j$ into HBRT, we compress its spatial information, retaining only temporal information, as the STAS task focuses on temporal modeling. The compressed feature $f_j^t$ will feed into $N_{1}$ hierarchical block Recurrent transformer block (HBRTB). Each HBRTB layer is consist of an dilated convolution, a BRT block with an dilated window mask, a feed-forward neural network and a gate neural network. The dilation rate of layer $o$ is set to $2^o$. The dilated convolution smooths the input feature $f_j^t$ \cite{farha2019ms}, while the feed-forward neural network improves the feature expression ability of the model. Gate neural network consist of activation functions and linear layers for selective memory and updating of historical information. Horizontal direction represents the current information flow, while its vertical direction represents the historical information flow in HBRT. In addition, we employ rotated relative position encoding \cite{su2021roformer} for each attention operation. Inspired by hierarchical representation design in ASformer \cite{yi2021asformer}, we modify hierarchical block attention operation of ASformer to a memory-friendly attention operation with dilated window mask. This modification not only enables the model to be trained on multiple samples but also improves its inference speed. Each layer's representation in HBRT is passed to its corresponding layer in next state, instead of BRT that only passed to the last layer or the last layer's preceding layer. This approach facilitates interaction of historical information when aggregating clustering features.

\subsubsection{Agent Model}
The agent model $\mathbb{A}$, parameterized by $\theta$, makes a decision $a_j$ based on current state $s_j$. As shown in Fig.\ref{model_fig}, the agent model is consist of a full-connect layer and $N_2$ dilated convolution blocks \cite{farha2019ms} which refine the result from full-connect layer. When evaluating the performance of the model, we will collect the decision $a_j$ made by the agent $\mathbb{A}$ and concatenate all the segmentation results from $a_0$ to $a_{\lceil \frac{T}{L} \rceil}$ in temporal order. Finally, evaluation indicators consistent with TAS tasks are adopted to ensure the substitution of STAS for TAS.

\subsubsection{Reward}

The formula is as follows:
\begin{align}
  r_j &= \beta_{1}^{\frac{1}{C} \sum_{c=0}^{C-1} \frac{2|a_{j,c} \cap a_{j,c}^*|}{|a_{j,c}|\cup |a_{j,c}^*|}} + \beta_{2} \\
  &= \beta_{1}^{\frac{1}{C} \sum_{c=1}^{C-1} \frac{2\sum_{i=j \cdot L}^{j \cdot L + k \cdot p} y_{i,c} p_{i,c}}{\sum_{i=j \cdot L}^{j \cdot L + k \cdot p}(y_{i,c} + p_{i,c})}} + \beta_{2}
\end{align}
where, $y_{i,c}$ represents the one-hot vector for class $c$ actions of frame $i$; $p_{i,c}$ represents the predicted probability from model for class $c$ actions of frame $i$; $\beta_1$ and $\beta_2$ is hyper-parameter.

An RL reward measures the value of decision $a_j$ which made by agent $\mathbb{A}$. In the SVTAS-RL model, this refers to the integrity of the action segment in the single-step decision of agent. We use a value function based on the dice coefficient \cite{milletari2016v} as a reward to measure it.

\subsection{Learning}

\subsubsection{Gradient Estimation}
\label{gradient_estimate}
The optimization objective and optimization direction of current STAS models are mismatched, which leads to the optimization dilemma. The essence of this phenomenon is that the gradient of the current supervised optimization method is unequal to the gradient of the optimization objective. Prove as follows:

The optimization objective of the STAS task is to maximize the action segment integrity of the entire video and we assume $Q(\cdot, \cdot)$ is a function which directly measure action segment integrity of video and $P(\cdot|I_j, \theta)$ means the distribution of $a_j$:
\begin{align}
\mathop{max}\limits_{\theta}\mathop{\mathbb{E}}\limits_{I\sim V}[\mathop{\mathbb{E}}\limits_{a_j\sim P(\cdot|I_j, \theta)}(Q(a, a^*))]
\end{align}
where, $V$ means video dataset, $I$ refers to image sequence corresponding to one video, $I_j$ is the image sequence corresponding to $j^{th}$ video clip, $\theta$ refers to the parameters of STAS model, $a$ is the prediction sequence corresponding to whole video, $a^*$ refers to label sequence corresponding to whole video, $j\in{0, \cdots q}$ is the video clip serial number, $q=\lceil \frac{T}{k \times p} \rceil $ is the max video clip serial number.

Since optimizing the entire video dataset is not practical, we consider maximizing expectation for a single video:

\begin{align}
  \mathbb{J}_{\theta} &= \mathop{max}\limits_{\theta}\mathop{\mathbb{E}}\limits_{a_j\sim P(\cdot|I_j, \theta)}(Q(a, a^*)) \\
  & \approx \mathop{max}\limits_{\theta}\mathop{\mathbb{E}}\limits_{a_j\sim P(\cdot|I_j, \theta)}(\sum_{j=0}^q Q(a_j, a^*_j))
  \label{expectation_equal}
\end{align}
where we approximate $Q(a, a^*)$ as $\sum_{j=0}^q Q(a_j, a^*_j)$, $a_j$ is the prediction sequence corresponding to $j^{th}$ video clip and $a^*_j$ is the label sequence corresponding to $j^{th}$ video clip.

Firstly, in order to use gradient descend algorithm to optimize parameters, we calculate the gradient of optimization function equation \ref{expectation_equal}. Then apply log derivative trick \cite{zhou2018deep} for it.

\begin{align}
  \nabla_{\theta}\mathbb{J}(\theta) &= \mathop{\mathbb{E}}\limits_{a_j\sim \pi_{\theta}(a_j|s_j)}[\sum_{j=0}^q Q(a_j, a^*_j) \nabla_{\theta} \log \pi_{\theta}(a_j|s_j)] \\
  &\approx \frac{1}{q}\sum_{j=0}^q Q(a_j,a^*_j) \nabla_{\theta} \log P(\cdot|I_j, \theta)
  \label{grad_equaltion}
\end{align}
where, $s_j$ refers to the state corresponding to $j^{th}$ video clip, $\pi_{\theta}(a_j|s_j)$ is the policy of the decision from model.

Secondly, since the probability space of $a_j$ is too large because of $a_j\in \mathbb{R}^{k\times C}$, and it's hard to compute directly, we approximate $P(\cdot|I_j, \theta)$ as $\frac{1}{k}\sum_{i=j\cdot L}^{j\cdot L + k\cdot p}P(l_{i,c}|I_j, \theta)$, where $c$ is prediction action index. So:

\begin{align}
  \label{cross_entropy}
  \nabla_{\theta}\mathbb{J}(\theta) &\approx \frac{1}{q} \sum_{j=0}^{q} Q(a_j,a^*_j) [ \frac{1}{k} \times \\
  & \qquad \qquad \sum_{i=j\cdot L}^{j\cdot L + k\cdot p} \nabla_{\theta} \log P(l_{i,c}|I_j, \theta)]  \nonumber \\
  & \approx \frac{1}{q} \sum_{j=0}^{q} Q(a_j,a^*_j) \frac{1}{k}\sum_{i=j\cdot L}^{j\cdot L + k\cdot p} [ \frac{1}{C} \times\\
  & \qquad \qquad \sum_{c=0}^{C-1} l^*_{i,c} \nabla_{\theta} \log P(l_{i,c}|I_j, \theta)]  \nonumber \\
  &=- \frac{1}{q} \sum_{j=0}^{q} Q(a_j,a^*_j) \nabla_{\theta} [ - \frac{1}{k \times C} \times\\
  & \qquad \qquad \sum_{i=j\cdot L}^{j\cdot L + k\cdot p} \sum_{c=0}^{C-1} l^*_{i,c} \log P(l_{i,c}|I_j, \theta)] \nonumber\\
  &=- \frac{1}{q} \sum_{j=0}^{q} Q(a_j,a^*_j) \nabla_{\theta} CE(a_j, a_j^*)
  \label{gradient_optimization_objective}
\end{align}
where $CE$ means cross entropy function. In order to facilitate the implementation, we approximate the Formula.\ref{cross_entropy} as the gradient form of cross entropy. And, gradient descent can be used by removing the negative sign in front of Formula.\ref{gradient_optimization_objective}. To sum up, we used many approximations, which made the estimated gradient biased. It will cause the center of the optimization direction deviated from the center of the optimization objective (See Fig.\ref{fig_op_dir_vs_op_obj}). But it still get better performance in optimizing STAS, because it can directly estimate the gradient of the optimization objective.

The optimization objective function of the supervised learning is to minimize the classification loss of a single frame:
\begin{align}
\mathop{min}\limits_{\theta}\mathbb{J}_{1}(\theta) &= - \mathop{min}\limits_{\theta}\frac{1}{q}\sum_{j=0}^{q} CE(a_j, a^*_j) \\
CE(a_j, a^*_j) &= - \frac{1}{k \times C} \sum_{i=j\cdot L}^{j\cdot L + k\cdot p} \sum_{c=0}^{C-1} l^*_{i,c} \log P(l_{i,c}|I_j, \theta) \\
\nabla_{\theta}\mathbb{J}_{1}(\theta) &= - \frac{1}{q} \nabla_{\theta} CE(a_j, a^*_j)
\label{supervised_optimization_objective}
\end{align}
where $l_{i,c}$ is the prediction probability of corresponding to $c^{th}$ action of $i^{th}$ image frame and $l^*_{i,c}$ is the one-hot label corresponding to $c^{th}$ action of $i^{th}$ image frame. It obviously only indirectly optimizes the action segment integrity of the entire video. Formula.\ref{supervised_optimization_objective} refers to the optimization gradient of supervised optimization method and it is difference from the gradient of STAS optimization objective Formula.\ref{gradient_optimization_objective}. There is a term of $\sum_{j=0}^{q} Q(a_j,a^*_j)$ difference between them.

\subsubsection{RL Optimization}
Inspired by RLHF, we introduce RL optimization algorithm into STAS task. It can use $\sum_{j=0}^{q} r_j$ as $\sum_{j=0}^{q} Q(a_j,a^*_j)$ for a accurate gradient estimation. The optimal action trajectory of a video is actually deterministic in our sequential decision-making task, which is denoted as $A^* = [a_0^*, a_1^*, \cdots, a_q]$, where $q = \lceil \frac{|V_n|}{k \times p} \rceil$. The optimization objective of the model is to maximize the accumulated rewards from each decision. The strategy learning methods for RL based on update methods are divided into temporal difference update and Monte Carlo update. So, we also designed two strategy learning methods, they are: Monte Carlo update learning method based on REINFORCE algorithm and temporal difference update learning method based on actor-critic algorithm. In RL tasks to maximize expectations, parameter updates typically use gradient ascent, while in computer vision tasks to minimize losses, parameter updates typically use gradient descent. In the STAS task, two parameter update algorithms that we designed using the same gradient descent algorithm as most computer vision tasks. We can repeat the process from Formula.\ref{expectation_equal} to Formula.\ref{gradient_optimization_objective} to prove that the gradient estimated by our optimization algorithm is equal to the gradient of optimization objective of the STAS task.

The Monte Carlo updated method usually uses the REINFORCE algorithm \cite{williams1992simple}. However, the original REINFORCE algorithm is updated using gradient ascent, so we modify the REINFORCE algorithm for the STAS task, following DSN \cite{zhou2018deep}. In our MC algorithm Algorithm.\ref{mc_training}, we use an approximate approach to estimate expectations. From an optimization perspective, the value function can be thought of as a variable coefficient that indicates how much confidence there is that the current gradient direction is the globally optimal gradient direction.

\begin{algorithm}[htbp]
  \SetKwData{Left}{left}\SetKwData{This}{this}\SetKwData{Up}{up}
  \SetKwFunction{Union}{Union}\SetKwFunction{FindCompress}{FindCompress}
  \SetKwInOut{Input}{input}\SetKwInOut{Output}{output}

  \Input{Agent model $\mathbb{A}(\cdot; \theta)$, Environment model $\mathbb{H}(\cdot, \cdot; \phi)$,
      Historical information $m_0$, Learning rate $\alpha$, Frame-skipping $p$, Frame-stacking $k$, Video set $V$}
  \Output{Sequence of action labels for each frame of untrimmed video, $[l_0, l_1, \cdots, l_T]$}
  \BlankLine
  Initial $\mathbb{A}(\cdot; \theta)$, $\mathbb{H}(\cdot, \cdot; \phi)$, $m_0$, result list $list$ \;
  \For{$n\leftarrow 0$ \KwTo $|V|$}{
    Sample $V_n$ from video set $V$\;
    \For{$j\leftarrow 0$ \KwTo $\lceil \frac{|V_n|}{k \times p} \rceil $}{
        Sample $I_j = [x_{j*L}, \cdots, x_{j*L + k*p} ]$ and $a_j^* = [l_{j*L}, \cdots, l_{j*L + k*p}]$ from Video $V_n$\;
        $s_j, m_{j+1} \leftarrow \mathbb{H}(I_j, m_j; \phi)$\;
        $a_j \leftarrow \mathbb{A}(s_j;\theta)$\;
        $r_j \leftarrow Q(a^*_j, a_j)$\;
        $list.append(a_j)$\;
    }
    $\mathbf{J}(\theta, \phi) = \mathbb{E}_{p_{\theta, \phi}(a0:\lceil \frac{|V_n|}{k \times p} \rceil)}[\sum_{j=0}^{\lceil \frac{|V_n|}{k \times p} \rceil}r_j]$\;
    \tcc{$P(\cdot|I_j, \theta, \alpha)$ is the distribution probability of $l_{i,c}$}
    $\nabla_{\theta, \phi}\mathbf{J}(\theta, \phi) = \frac{1}{q} \sum_{j=0}^{q} r_j \nabla_{\theta, \alpha} CrossEntropy(a_j, a_j^*)$\;
    $\{\theta, \phi\} \leftarrow \{\theta, \phi\} - \alpha \nabla_{\theta, \phi}\mathbf{J}(\theta, \phi)$\;
  }
  \caption{Monte Carlo Episodic REINFORCE Learning for STAS task (MC)}\label{mc_training}
\end{algorithm}

\begin{algorithm}[htbp]
  \SetKwData{Left}{left}\SetKwData{This}{this}\SetKwData{Up}{up}
  \SetKwFunction{Union}{Union}\SetKwFunction{FindCompress}{FindCompress}
  \SetKwInOut{Input}{input}\SetKwInOut{Output}{output}

  \Input{Agent model $\mathbb{A}(\cdot; \theta)$, Environment model $\mathbb{H}(\cdot, \cdot; \phi)$,
      Historical information $m_0$, Learning rate $\alpha$, Frame-skipping $p$, Frame-stacking $k$, Video set $V$}
  \Output{Sequence of action labels for each frame of untrimmed video, $[l_0, l_1, \cdots, l_T]$}
  \BlankLine
  Initial $\mathbb{A}(\cdot; \theta)$, $\mathbb{H}(\cdot, \cdot; \phi)$, $m_0$, result list $list$ \;
  \For{$n\leftarrow 0$ \KwTo $|V|$}{
    Sample $V_n$ from video set $V$\;
    \For{$j\leftarrow 0$ \KwTo $\lceil \frac{|V_n|}{k \times p} \rceil $}{
        Sample $I_j = [x_{j*L}, \cdots, x_{j*L + k*p} ]$ and $a_j^* = [l_{j*L}, \cdots, l_{j*L + k*p}]$ from Video $V_n$\;
        $s_j, m_{j+1} \leftarrow \mathbb{H}(I_j, m_j; \phi)$\;
        $a_j \leftarrow \mathbb{A}(s_j;\theta)$\;
        $r_j \leftarrow Q(a^*_j, a_j)$\;
        $list.append(a_j)$\;
        $\mathbf{J}(\theta, \phi) = r_j$\;
        $\nabla_{\theta, \phi}\mathbf{J}(\theta, \phi) = r_j \nabla_{\theta, \alpha} CrossEntropy(a_j, a_j^*)$\;
        $\{\theta, \phi\} \leftarrow \{\theta, \phi\} - \alpha \nabla_{\theta, \phi}\mathbf{J}(\theta, \phi)$\;
    }
  }
  \caption{Temporal Difference Actor-Critic Learning for STAS task (TD)}\label{td_training}
\end{algorithm}

\begin{figure*}[htbp]
  \centering
  \includegraphics[width=\linewidth]{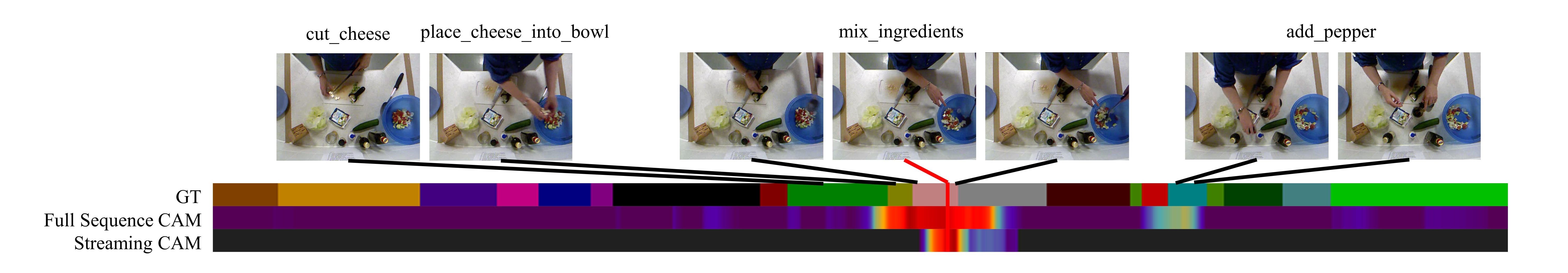}
  \caption{The importance of global contextual information. Class Activation Map (CAM) for red frame in full sequence TAS and STAS. Streaming CAM is mostly black because the frames are not available at inference time.}
  \label{context}
\end{figure*}

\begin{table*}[htbp]
  \caption{Comparison of clustering feature and sequential feature. Sequence means sequential feature and cluster means clustering feature.\label{compare_clustering}}
  \centering
  \begin{tabular}{@{}cccclllll@{}}
    \toprule
      dataset & paradigm & model & modality & Acc & Edit & F1@0.1 & F1@0.25 & F1@0.5 \\ \midrule
      Breakfast & sequence & HBRT & rgb & 49.3 & 56.4 & 55.6 & 48.9 & 34.6 \\ 
      Breakfast & sequence & HBRT & flow & 61.8 & 65.3 & 66.5 & 60.9 & 47.6 \\ 
      Breakfast & sequence & HBRT & rgb+flow & 62.9 & 67.9 & 68.1 & 62.7 & 48.7 \\
      Breakfast & cluster & SVTAS(ours) & rgb & \textbf{65.6} & \textbf{70.9} & \textbf{71.3} & \textbf{64.9} & \textbf{49.8} \\
      \midrule
      50salads  & sequence & ASformer & rgb & 63.4 & 47.4 & 52.9 & 48.7 & 38.2 \\
      50salads  & cluster & ASformer & rgb & \textbf{77.9} & \textbf{68.2} & \textbf{75.7} & \textbf{73.2} & \textbf{64.4} \\
      \midrule
      50salads  & sequence & FC & rgb & 53.6 & 5.5 & 7.0 & 4.2 & 2.4 \\
      50salads  & cluster & FC & rgb & \textbf{73.1} & \textbf{34.1} & \textbf{43.9} & \textbf{40.3} & \textbf{33.2} \\
      \midrule
      gtea      & sequence & ASformer & rgb & 68.9 & 70.2	& 76.1 & 72.6 & 60.8\\
      gtea      & cluster & ASformer & rgb & \textbf{73.7} & \textbf{81.3} & \textbf{86.2} & \textbf{83.3} & \textbf{72.3} \\
      \midrule
      gtea      & sequence & FC & rgb & 53.4 & 30.0 & 33.1 & 26.2 & 18.1 \\
      gtea      & cluster & FC & rgb & \textbf{64.0} & \textbf{45.9} & \textbf{51.8} & \textbf{46.9} & \textbf{37.7} \\
    \bottomrule
  \end{tabular}
\end{table*}

The actor-critic algorithm \cite{sutton1999policy} is a common algorithm for the temporal difference updated method. As with the original REINFORCE algorithm, we also modify it to a gradient descent version Algorithm.\ref{td_training}, and since our critic model can be estimated directly by the algorithm without bias, we only need to update the parameters of the agent in a single step. Note that we directly use the reward here as the error of the temporal difference, which is crude, but also allows the model to be optimized roughly toward the optimization objective of the STAS task.

\section{Experiment and Discussion}
\subsection{Datasets and Evaluation Metrics}

\textbf{Datesets}: The \textbf{GTEA} \cite{fathi2011learning} dataset contains 28 videos corresponding to 7 different activities, performed by 4 subjects. On average, there are 20 action instances per video. The evaluation was performed by excluding one subject to use cross-validation. The \textbf{50Salads} \cite{stein2013combining} dataset contains 50 videos with 17 action classes. On average, each video contains 20 action instances. 50Salads also uses five-fold cross-validation. The \textbf{Braekfast} \cite{kuehne2014language} dataset is the largest of the four datasets. In total, Braekfast contains 77 hours of 1712 videos. It contains 48 different actions, and each video contains an average of 6 action instances. Also, it will be evaluated by four-fold cross-validation. The \textbf{EGTEA} \cite{li2018eye} dataset has the longest average video length in the four datasets. In total, EGTEA contains 28 hours of cooking activities from 86 unique sessions of 32 subjects. It contains 20 different actions, and each video contains an average of 45 action instances. Also, it will be evaluated by three-fold cross-validation.

\textbf{Metrics}: To evaluate STAS task results, we adopt several metrics including frame-wise accuracy (\textbf{Acc}) \cite{farha2019ms}, segmental edit distance (\textbf{Edit}) \cite{farha2019ms}, and the F1 score at temporal IoU threshold 0.1, 0.25, 0.5 (denote by \textbf{F1@\{0.1, 0.25, 0.5\}}), \cite{li2022bridge}. F1 score is proposed to measure the integrity of action segment and \textbf{F1@0.5} is the most \textbf{important} indicator for TAS. Edit score is measure the action sequence distance between the inferred result and the ground true. The frame-wise accuracy is measure quality of single-frame classification.

\subsection{Implementation Details}
We adopt AdamW optimizer and the base learning rate of $5 \times 10^{-4}$ with a $1 \times 10^{-4}$ weight decay. The spatial resolution of the input video is $224 \times 224$. We use Kinetics-600 \cite{carreira2017quo} pre-trained weight for all feature extractors. The $k$x$p$ for GTEA is 64x2, for 50Salads, Breakfast and EGTEA is 128x8. The model is trained 80 epochs with batch size is set to 1 for GTEA and 50Salads and trained 50 epochs with batch size is set to 1 for Breakfast and EGTEA. $\beta_1$ is 4 and $\beta_2$ is -1. $D$ is set to 128 and $M$ is set to 512.

\subsection{Impact of TAS Migration to STAS}

In Tab.\ref{migration_experiment}, we observe that migrating TAS models to STAS generates huge performance gap, and even HBRT designed for streaming data still cannot fill the performance gap, which indicates that changing TAS into STAS is a challenging task. A closer look reveals that optical flow modality with temporal information play an important role in TAS, but to achieve end-to-end segmentation we will only use rgb modality, which makes the end-to-end STAS more challenging. Fig.\ref{context} shows that in full sequence TAS, global contextual information is required. However, in streaming scenario, the model will rely entirely on the information of the current video clip, which shows that TAS models cannot be migrated directly to STAS task.

\begin{table}[htbp]
  \caption{Migration experiment from TAS to STAS in Breakfast. \dag means migration experiment.}
  \centering
  \scalebox{.7}{
    \begin{tabular}{@{}cllllllll@{}}
    \toprule
    \multicolumn{1}{l}{}       & Model    & paradigm & modality       & Acc  & Edit & \multicolumn{3}{c}{F1@\{0.1,0.25,0.5\}} \\ \midrule
    \multirow{2}{*}{\rotatebox{90}{full}}      & ASformer & sequence & rgb feature      & 56.3 & 63.2 & 63.6        & 56.5        & 41.2        \\
    & ASformer & sequence & rgb+flow feature & \textbf{73.5} & \textbf{75.0} & \textbf{76.0} & \textbf{70.6} & \textbf{57.4} \\ \midrule
    \multirow{5}{*}{\rotatebox{90}{streaming}} & ASformer\dag & sequence & rgb+flow feature & 52.7 & 57.2 & 51.3        & 51.3        & 37.7        \\
                              & HBRT     & sequence & rgb feature      & 49.3 & 56.4 & 55.6        & 48.9        & 34.6        \\
                              & HBRT     & sequence & flow feature     & 61.8 & 65.3 & 66.5        & 60.9        & 47.6        \\
    & HBRT     & sequence & rgb+flow feature & \textbf{62.9} & \textbf{67.9} & \textbf{68.1} & \textbf{62.7} & \textbf{48.7} \\ \cmidrule(l){2-9} 
                              & SVTAS(ours)  & cluster  & rgb              & 65.6 & 70.9 & 71.3        & 64.9        & 49.8        \\ \bottomrule
    \end{tabular}
  }
  \label{migration_experiment}
\end{table}

\begin{table*}[htbp]
  \caption{Comparison with the state-of-the-art results on four datasets. Global action segment integrity was measured by \textbf{F1} metrics. \textbf{Bold} and \underline{underlined} denote the best and second-best results in each column, respectively. \dag means migration experiment. Streaming feature is rgb + flow (optical flow) features, so it is \textbf{unfair} comparison when across horizontal line. But we only use rgb modality to achieve comparable results of the full sequence by end-to-end streaming method.}
  \centering
  \scalebox{0.68}{
  \begin{tabular}{@{}ccl|lllll|lllll|lllll|lllll@{}}
  \toprule
  \multicolumn{1}{l}{}       & \multicolumn{1}{l}{}                          & Dataset            & \multicolumn{5}{c|}{GTEA}                                                     & \multicolumn{5}{c|}{50Salads}                                                 & \multicolumn{5}{c|}{Berakfast}                                                & \multicolumn{5}{c}{EGTEA}                                                     \\ \cmidrule(l){4-23} 
  \multicolumn{1}{l}{}       & \multicolumn{1}{l}{}                          & Metric             & Acc           & Edit          & \multicolumn{3}{c|}{F1@\{0.1,0.25,0.5\}}      & Acc           & Edit          & \multicolumn{3}{c|}{F1@\{0.1,0.25,0.5\}}      & Acc           & Edit          & \multicolumn{3}{c|}{F1@\{0.1,0.25,0.5\}}      & Acc           & Edit          & \multicolumn{3}{c}{F1@\{0.1,0.25,0.5\}}       \\ \midrule
  \multicolumn{2}{c|}{\multirow{16}{*}{\rotatebox{90}{full (rgb + flow feature)}}}
  & Bi-LSTM \cite{singh2016multi} & 55.5          & -             & 66.5          & 59.0          & 43.6          & 55.7          & 55.6          & 62.6          & 58.3          & 47.0          & -             & -             & -             & -             & -             & 70            & 28.5          & 27            & 23.1          & 15.1          \\
  \multicolumn{2}{c|}{}
  & Dilated TCN \cite{lea2017temporal} & 58.3          & -             & 58.8          & 52.2          & 42.2          & 59.3          & 43.1          & 52.2          & 47.6          & 37.4          & -             & -             & -             & -             & -             & -             & -             & -             & -             & -             \\
  \multicolumn{2}{c|}{}
  & ST-CNN \cite{lea2016segmental} & 60.6          & -             & 58.7          & 54.4          & 41.9          & 59.4          & 45.9          & 55.9          & 49.6          & 37.1          & -             & -             & -             & -             & -             & -             & -             & -             & -             & -             \\
  \multicolumn{2}{c|}{}
  & ED-TCN \cite{lea2017temporal} & 64.0          & -             & 72.2          & 69.3          & 56.0          & 64.7          & 52.6          & 68.0          & 63.9          & 52.6          & 43.3          & -             & -             & -             & -             & \textbf{70.1} & 28.6          & 31.1          & 27.7          & {\ul 19.6}    \\
  \multicolumn{2}{c|}{}
  & TDRN \cite{lei2018temporal} & 70.1          & 74.1          & 79.2          & 74.4          & 62.7          & 68.1          & 66.0          & 72.9          & 68.5          & 57.2          & -             & -             & -             & -             & -             & -             & -             & -             & -             & -             \\
  \multicolumn{2}{c|}{}
  & MS-TCN \cite{farha2019ms} & 76.3          & 79.0          & 85.8          & 83.4          & 69.8          & 80.7          & 67.9          & 76.3          & 74.0          & 64.5          & 66.3          & 61.7          & 52.6          & 48.1          & 37.9          & 69.2          & {\ul 32.2}    & {\ul 32.1}    & {\ul 28.3}    & 18.9          \\
  \multicolumn{2}{c|}{}
  & MS-TCN++ \cite{li2020ms} & 80.1          & 83.5          & 88.8          & 85.7          & 76.0          & 83.7          & 74.3          & 80.7          & 78.5          & 70.1          & 67.6          & 65.6          & 64.1          & 58.6          & 45.9          & -             & -             & -             & -             & -             \\
  \multicolumn{2}{c|}{}
  & BCN \cite{BCN} & 79.8          & 84.4          & 88.5          & 87.1          & 77.3          & 84.4          & 74.3          & 82.3          & 81.3          & 74.0          & 70.4          & 66.2          & 68.7          & 65.5          & 55.0          & -             & -             & -             & -             & -             \\
  \multicolumn{2}{c|}{}
  & Global2Local \cite{gao2021global2local} & 78.5          & 84.6          & 89.9          & 87.3          & 75.8          & 82.2          & 73.4          & 80.3          & 78.0          & 69.8          & 70.7          & 73.3          & 74.9          & 69.0          & 55.2          & -             & -             & -             & -             & -             \\
  \multicolumn{2}{c|}{}
  & ASRF \cite{asrf} & 77.3          & 83.7          & 89.4          & 87.8          & 79.8          & 84.5          & 79.3          & 84.9          & 83.5          & 77.3          & 67.6          & 72.4          & 74.3          & 68.9          & 56.1          & -             & -             & -             & -             & -             \\
  \multicolumn{2}{c|}{}
  & C2F-TCN \cite{singhania2021coarse} & 80.8          & 86.4          & 90.3          & 88.8          & 77.7          & 84.9          & 76.4          & 84.3          & 81.8          & 72.6          & 76.0          & 69.6          & 72.2          & 68.7          & 57.6    & -             & -             & -             & -             & -             \\
  \multicolumn{2}{c|}{}
  & ASFormer \cite{yi2021asformer} & 79.7          & 84.6          & 90.1          & 88.8          & 79.2          & 85.6          & 79.6          & 85.1          & 83.4          & 76.0          & 73.5    &  {\ul 75.0}  &  76.0    &  70.6    & 57.4          & -             & -             & -             & -             & -             \\
  \multicolumn{2}{c|}{}
  & m-GRU+GTRM \cite{huang2020improving} & -             & -    & -    & -    & -    & -             & -    & -    & -             & -             & -             & -             & -             & -             & -             & {\ul 69.5}    & \textbf{41.8} & \textbf{41.6} & \textbf{37.5} & \textbf{25.9} \\
  \multicolumn{2}{c|}{}
  & bridge-prompt \cite{li2022bridge} & {\ul 81.2} & {\ul 91.6}    & \textbf{94.1} & \textbf{92.0} & {\ul 83.0} & {\ul 88.1} &  83.8    & {\ul 89.2} & {\ul 87.8} &  81.3     & -             & -    & -    & -    & -             & -             & -             & -             & -             & -             \\
  \multicolumn{2}{c|}{}
  & UVAST \cite{behrmann2022unified} & 80.2     & \textbf{92.1} & {\ul 92.7}    & 91.3    &  81.0    &  87.4    & {\ul 83.9} &  89.1    &  87.6     & {\ul 81.7} & \textbf{77.1} & 69.7    & {\ul 76.9} & {\ul 71.5} & {\ul 58.0} & -             & -             & -             & -             & -             \\
  \multicolumn{2}{c|}{}
  & DiffAct \cite{liu2023diffusion} & \textbf{82.2}    & 89.6 & 92.5    & {\ul 91.5}    & \textbf{84.7}    & \textbf{88.9}    & \textbf{85.0} & \textbf{90.1}    & \textbf{89.2}    & \textbf{83.7} & {\ul 76.4} & \textbf{78.4}    & \textbf{80.3} & \textbf{75.9} & \textbf{64.6} & -             & -             & -             & -             & -             \\\midrule
  \multirow{10}{*}{\rotatebox{90}{streaming}} & \multicolumn{1}{c|}{\multirow{6}{*}{\rotatebox{90}{feature (rgb+flow)}}}
  & TOT+TCL \cite{kumar2022unsupervised} & -   & -   & -   & -   & -  & - & - & - & - & -  & -  & -  & -  & - & 25.1  & -   & -    & -    & -   & -    \\
  & \multicolumn{1}{c|}{}
  & OAS \cite{ghoddoosian2022weakly} & -   & -   & -   & -   & -  & -   & -   & -   & -   & -  & 41.6  & -  & -  & - & -  & -   & -    & -    & -   & -    \\
  & \multicolumn{1}{c|}{}
  & IDT+LM \cite{richard2016temporal} & -   & -   & -   & -   & -  & 48.7 & 45.8 & 44.4 & 38.9 & 27.8  & -  & -  & -  & - & -  & -   & -    & -    & -   & -    \\
  & \multicolumn{1}{c|}{}
  & ASformer \cite{yi2021asformer} \dag & \textbf{76.3} & \textbf{79.7} & \textbf{85.4} & \textbf{83.1} & \textbf{72.8} & 70.0 & 54.0 & 62.1 & 57.6 & 49.1 & 52.7 & 57.2 & 57.7 & 51.3 & 37.7 & -  &  -  &   -  &  -  &  -   \\
  & \multicolumn{1}{c|}{}
  & DiffAct \cite{liu2023diffusion} \dag & 58.7 & 59.6 & 66.0 & 61.8 & 48.9 & 40.5 & 30.2 & 32.3 & 29.0 & 19.8 & 45.7 & 48.2 & 46.2 & 40.9 & 29.6  & -   & -    & -    & -   & -    \\
  & \multicolumn{1}{c|}{}
  & HBRT(ours)        & 74.9 & 78.7 & 84.3 & 82.2 & 72.5 & \textbf{74.2} & \textbf{56.2} & \textbf{63.7} & \textbf{60.4} & \textbf{52.1} & \textbf{62.9} & \textbf{67.9} & \textbf{68.1} & \textbf{62.7} & \textbf{48.7} &  -  & - &   -  &  -  &  -   \\ \cmidrule(l){4-23} 
  & \multicolumn{1}{c|}{\multirow{4}{*}{\rotatebox{90}{video(rgb)}}}
  & ETSN \cite{kang2022efficient} & 78.3          & 79.9          & 87.1          & 84.5          & 71.8          & 83.1          & 71.1          & 79.0          & 76.8          & 69.5          & -             & -             & -             & -             & -             & -             & -             & -             & -             & -             \\
  & \multicolumn{1}{c|}{}
  & SVTAS(ours)        & 79.5 & 83.5 & 88.7 & 86.2 & 77.6 & 86.7 & 78.4 & 85.3 & 83.7 & 77.2 & \textbf{65.6} & \textbf{70.9} & \textbf{71.3} & 64.9 & \textbf{49.8} & 69.6 & 47.3 & 50.1 & 46.0 & \textbf{32.8}              \\
  & \multicolumn{1}{c|}{}
  & SVTAS-RL(TD)(ours) & \textbf{79.9} & \textbf{86.4} & \textbf{90.9} & \textbf{88.7} & 80.0 & \textbf{87.4} & \textbf{79.8} & \textbf{86.1} & \textbf{85.0} & 79.6 & 64.9 & 70.6 & \textbf{71.3} & \textbf{65.0} & 49.4 & \textbf{69.7} & \textbf{47.9} & \textbf{51.2} & \textbf{46.9} & \textbf{32.8} \\
  & \multicolumn{1}{c|}{}
  & SVTAS-RL(MC)(ours) & 78.8 & \textbf{86.4} & 90.0 & 88.4 & \textbf{81.1} & 87.3 & 78.9 & 85.9 & 84.2 & \textbf{80.1} & 63.7 & 70.1 & 70.5 & 64.3 & 49.6 & 68.8 & 47.4 & 49.8 & 45.4 & 32.2 \\ \bottomrule
  \end{tabular}
  }
  \label{compare_sota}
\end{table*}

\begin{figure}[htbp]
  \centering
  \includegraphics[width=\linewidth]{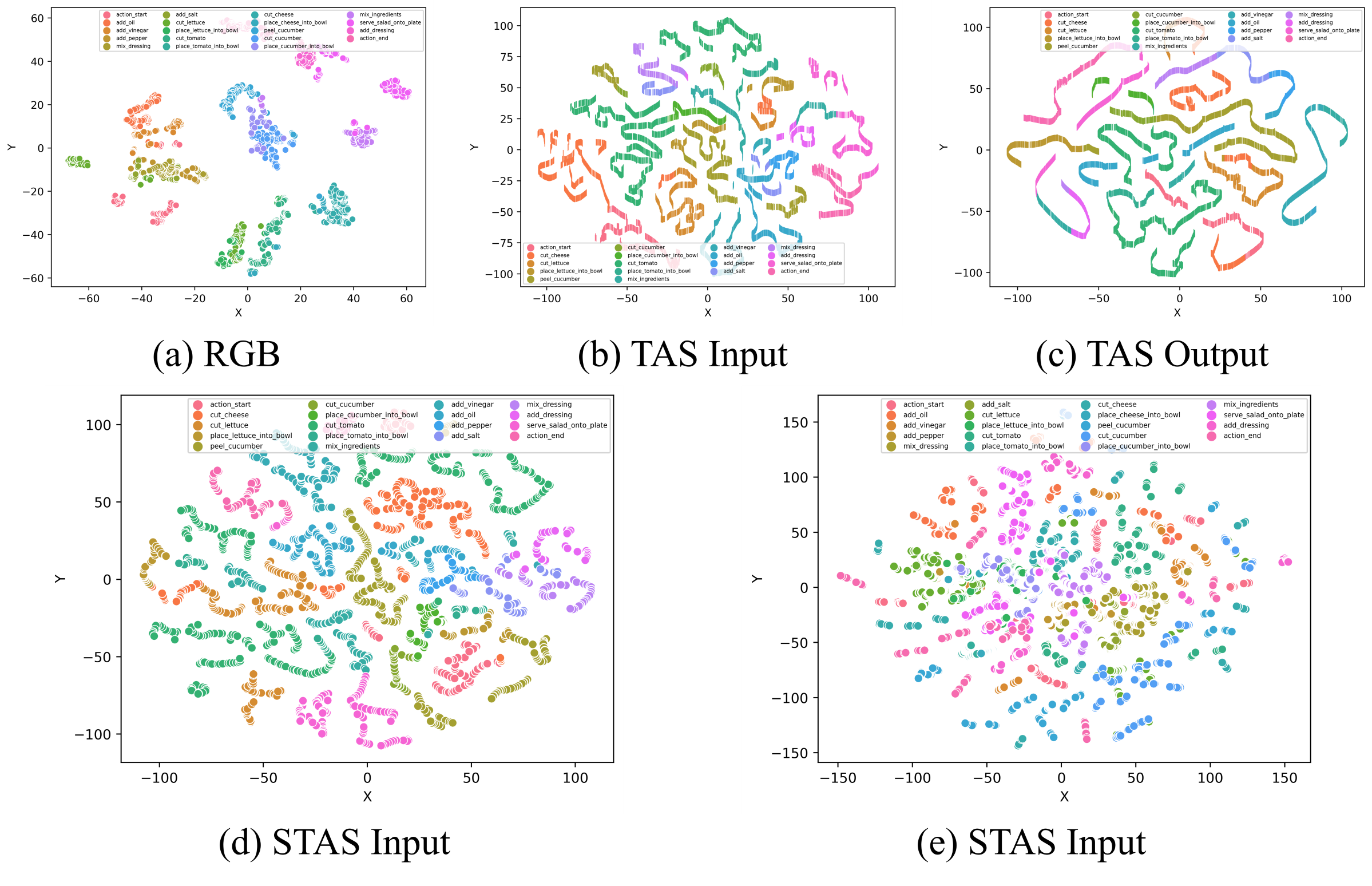}
  \caption{Comparison of feature manifold. All features are visualized by t-SNE. Obviously, TAS is a sequence-to-sequence transformation paradigm and STAS is a clustering paradigm.}
  \label{t_sne_feature}
\end{figure}

\subsubsection{Modeling Bias}
As can be seen in Fig.\ref{t_sne_feature}, TAS uses sequential features extracted through a sliding window, which is a tangled line in the feature manifold. It is suitable for sequence-to-sequence transformation task. The existence of modeling bias makes TAS models perform poorly on STAS task. And, as shown in Tab.\ref{compare_clustering}, it is clear that the segmentation results of HBRT, which did not use the clustering feature (Line.2), are significantly lower in various modalities compared to SVTAS (Line.5) using only the rgb modality. This effectively demonstrates the positive enhancement of the clustering feature for the performance of HBRT. To further prove the importance of the clustering feature for STAS tasks, we conducted experiments on ASformer \cite{yi2021asformer} model and FullConnect model on the 50salads and GTEA dataset. Both ASformer and FC with clustering features achieved significantly improved segmentation results compared to those using sequential features. We believe this effectively proves that clustering features not only have a positive impact on our designed HBRT, but are also extremely important for STAS task.

\begin{figure}[htbp]
  \centering
  \includegraphics[width=\linewidth]{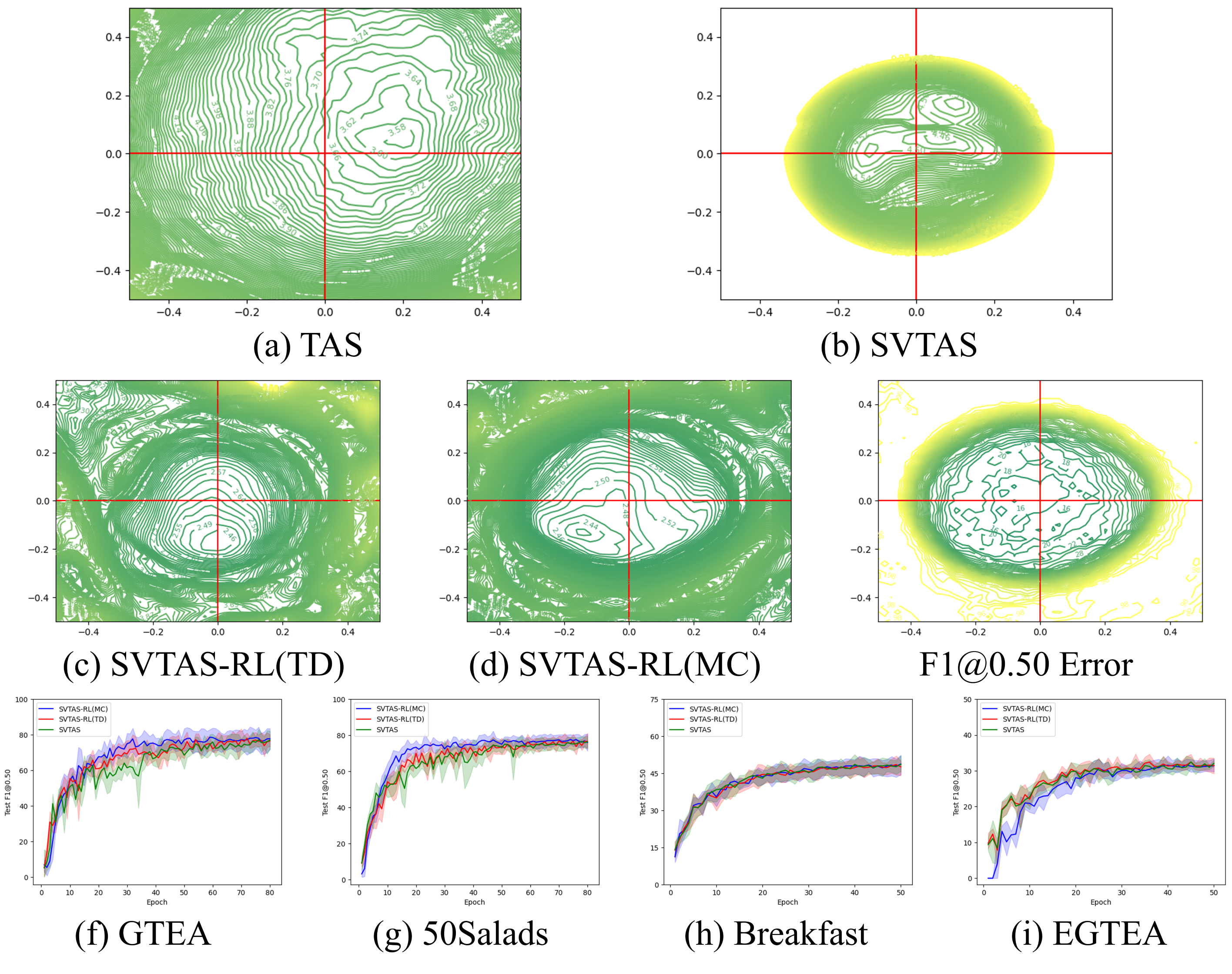}
  \caption{Visualize optimization objective and optimization direction by 2D surface. (a)-(d) are optimization direction surface. (e) is optimization objective surface. (f)-(i) is training state.}
  \label{fig_op_dir_vs_op_obj}
\end{figure}

\begin{table}[htbp]
  \caption{Comparison of training method and feature manifold on STAS in split1 of 50Salads. sep means training separately. e2e means end-to-end training. \dag means pre-trained VE from an end-to-end SVTAS. FC means full connect layer.}%
  \centering
    \begin{tabular}{@{}ccllllll@{}}
    \toprule
    \multicolumn{1}{l}{} & Model & Acc           & Edit          & \multicolumn{3}{c}{F1@\{0.1,0.25,0.5\}}       \\ \midrule
    \multirow{2}{*}{e2e} & VE\dag+FC & 81.1          & 50.1          & 63.0          & 59.1          & 50.1          \\
    & VE+FC    & \textbf{82.6} & \textbf{60.7} & \textbf{66.7} & \textbf{64.4} & \textbf{54.9} \\ \midrule
    sep                  & SVTAS & 84.4          & 75.4          & 83.6          & \textbf{82.7} & 72.9          \\
    e2e                  & SVTAS & \textbf{85.1} & \textbf{76.0} & \textbf{83.9} & 82.5          & \textbf{74.4} \\ \bottomrule
    \end{tabular}
  \label{compare_e2e}
\end{table}

\subsubsection{Optimization Dilemma}
Fig.\ref{fig_op_dir_vs_op_obj} (e) shows that (0, 0) is the center of the optimization objective. The optimization direction surfaces of models all have different degrees offset to the optimization objective center, which means that neither the previous nor our proposed method can unify the optimization objective and the optimization direction. However, our method with RL can maintain convexity near the center of the optimization objective, enabling the model to achieve a global optimum in the optimization direction. The optimization surface of the model without RL has many local optima that are difficult to optimize. Fig.\ref{fig_op_dir_vs_op_obj} (f)-(i) proves that the optimization method with RL we propose can update the model parameters faster and better during the training process.

\subsubsection{Comparison between end-to-end training and training separately}
We consider two types of training steps: end-to-end training and training separately. Similar to TAS, training separately is possible for VE and temporal models. Tab.\ref{compare_e2e} shows that end-to-end training is better than training separately.

\subsection{Comparison to prior work}

\begin{figure}[htbp]
  \centering
  \includegraphics[width=\linewidth]{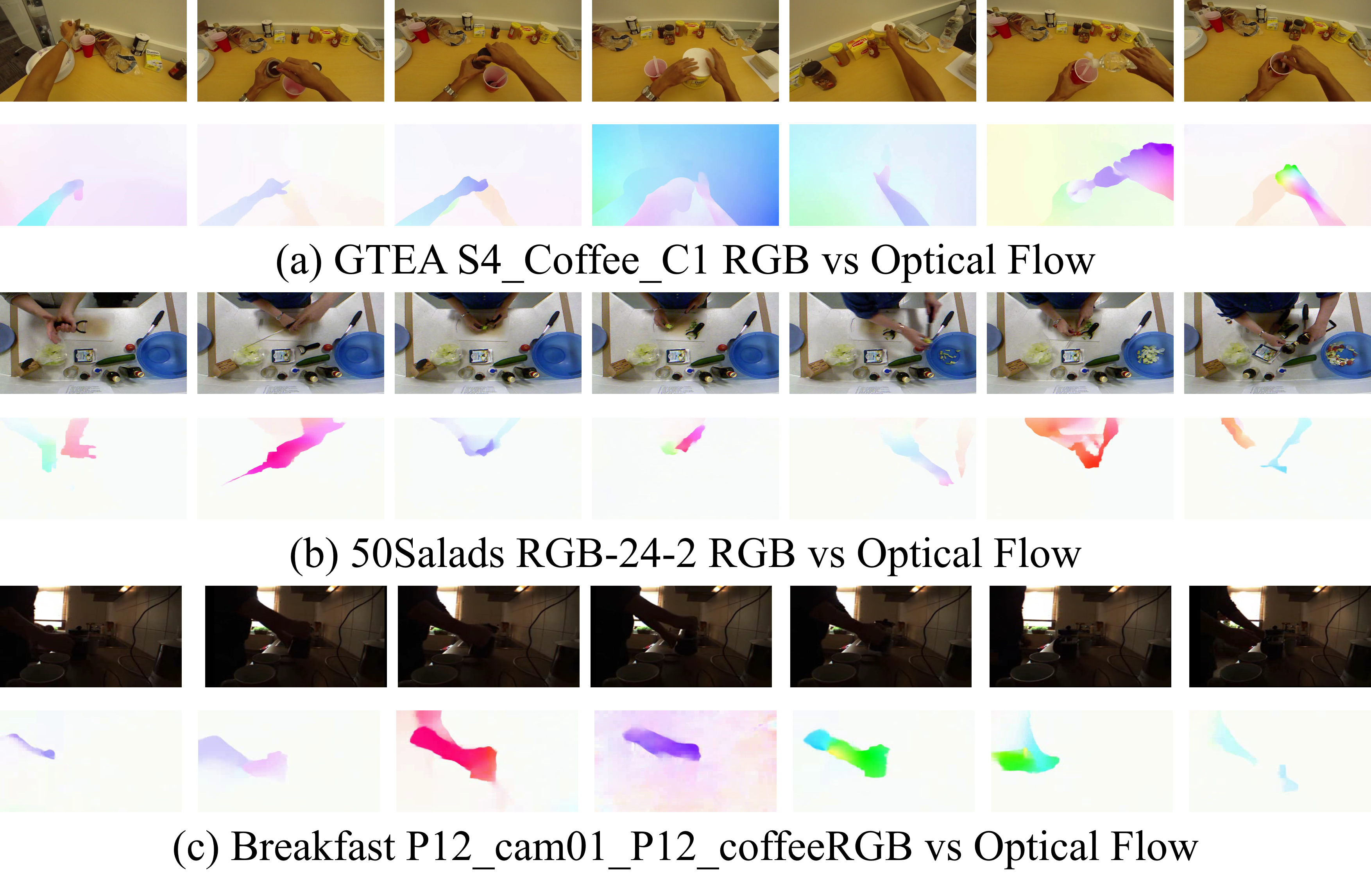}
  \caption{RGB Modality vs Optical Flow Modality.}
  \label{dataset_modality_comp}
\end{figure}

We show in Tab.\ref{compare_sota} the comparison results of two segmentation paradigms: TAS and STAS. We can observe that the end-to-end SVTAS approaches are already very comparable to the SOTA model of the current TAS model. And our method even outperforms the latter on the dataset EGTEA, which indicates that the stream-based approach is better than latter for ultra-long videos action segmentation. Although the SVTAS-RL(MC) approach is lower than the SVTAS-RL(TD) on F1@0.1 and F1@0.25, the former performs better on F1@0.5. Just as the metric in object detection uses an IOU threshold of 0.5 as a more important benchmark for comparison, indicating that the former model is more accurate in integrity of action segment through guidance from RL reward. In Tab.\ref{compare_sota}, Breakfast does not perform as expected. We believe that this is caused by the poor quality of the rgb modality for Breakfast dataset. As shown in Fig.\ref{dataset_modality_comp}, we show some samples from GTEA, 50Salads and Breakfast, and compare their RGB modality and optical flow modality. We can observe that the RGB modality of GTEA and 50Saldas has clear object boundaries. However, it is difficult to distinguish object boundaries even with human eyes in RGB modality of Breakfast. In the optical flow modality, because it will be extracted through the optical flow model, even Breakfast can have good object boundaries and filter out a lot of irrelevant information, which can improve the discriminability of actions \cite{sevilla2019integration}. Existing TAS models are mostly multi-modality models that take both RGB modality and optical flow modality as input. This means that even if the RGB modality of samples in Breakfast is extremely poor, the required feature information can still be extracted from optical flow data (See Tab.\ref{compare_clustering}). However, It is an end-to-end model that our designed SVTAS-RL is, which only takes RGB data as input. This makes the model perform poorly on the Breakfast dataset. If the same RGB modality is used, our proposed model has already exceeded the performance of the full sequence (See Tab.\ref{compare_clustering}).

\subsection{Ablation Study}
\subsubsection{Duration of Video Clip}
The experiments in Tab.\ref{k_p_ablation} are all on split1 and EGTEA Avg.$T$ is 28157.7. From Tab.\ref{k_p_ablation} we can observe that there are three principles for the selection of $k$ and $p$: (a) The larger $k$ is in a certain range, the better within limits. It is consistent with the modeling bias we observed in Fig.\ref{model_bias}. (b) The selection of $p$ is related to the current dataset and its effect is not as significant as $k$. (c) The combination of $k$ and $p$ is related to the average number of frames per video in the current dataset. Overall, SVTAS-RL method we propose is very suitable for the data modality under the STAS task.

\begin{table}[htbp]
  \caption{Ablation experiment of $k$ and $p$. Avg.$T$ means average number of frames per video.}
  \centering
  \scalebox{0.8}{
  \begin{tabular}{@{}ccllllll@{}}
  \toprule
  \multicolumn{1}{l}{Avg.$T$} & \multicolumn{1}{l}{Dataset} & $k$ x $p$ & Acc & Edit & \multicolumn{3}{c}{F1@\{0.1,0.25,0.5\}} \\ \midrule
  \multirow{4}{*}{1115.2}  & \multirow{4}{*}{GTEA}      & 16 x 2   & 75.2          & 69.7          & 79.7          & 75.7          & 67.7          \\
                           &                            & 32 x 2   & 77.8          & 80.6          & 84.4          & 81.0          & 72.7          \\
                           &                            & 64 x 2   & \textbf{79.9} & 85.6          & \textbf{88.0} & \textbf{86.5} & \textbf{79.3} \\
                           &                            & 128 x 2  & 79.2          & \textbf{85.7} & 87.7          & 84.8          & 78.1          \\ \midrule
  \multirow{4}{*}{11551.9} & \multirow{4}{*}{50Salads}  & 128 x 2  & 84.1          & 60.2          & 67.9          & 65.8          & 57.4          \\
                           &                            & 128 x 4  & 84.6          & 70.8          & 79.0          & 75.6          & 69.6          \\
                           &                            & 128 x 8  & \textbf{85.7} & 75.7          & 83.6          & 82.3          & \textbf{76.4} \\
                           &                            & 128 x 12 & 85.3          & \textbf{77.4} & \textbf{85.8} & \textbf{84.3} & 75.7          \\ \midrule
  \multirow{4}{*}{2097.5}  & \multirow{4}{*}{Breakfast} & 32 x 32  & 57.8          & 62.5          & 64.1          & 58.6          & 37.2          \\
                           &                            & 64 x 16  & 62.1          & 65.9          & 66.5          & 60.8          & 44.6          \\
                           &                            & 128 x 8  & 63.2          & \textbf{68.8} & \textbf{68.7} & 63.2          & 47.4          \\
                           &                            & 256 x 4  & \textbf{64.3} & 66.8          & \textbf{68.7} & \textbf{63.5} & \textbf{49.7} \\ \bottomrule
  \end{tabular}
  }
  \label{k_p_ablation}
\end{table}

\subsubsection{Architecture of HBRT}
Tab.\ref{hbrt_ablation} shows the ablation experiments on the HBRT structure. We can observe that the addition of memory information in the vertical direction enhances the Edit score. This indicates that the network is able to model past action information through the extraction and updating of memory information. And it enhances the model ability to infer sequential actions. Line.4 show that HBRT can enhance integrity of action segment through the passing hierarchical memory information.

\begin{table}[htbp]
  \caption{vertical and horizontal direction ablation experiments about HBRT in the feature modality of GTEA with HBRT. -2 means only pass last but not least layer memory to next state.}
  \centering
  \begin{tabular}{@{}lllllll@{}}
  \toprule
  model             & pass       & Acc  & Edit & \multicolumn{3}{c}{F1@\{0.1,0.25,0.5\}} \\ \midrule
  horizontal attn     & -          & 74.1 & 75.6 & 82.9 & 81.4 & 69.6 \\
  + vertical attn & -2         & 74.5 & \textbf{79.7} & 83.7 & 81.3 & 71.0 \\
  HBRT              & all        & \textbf{74.9} & 78.7 & \textbf{84.3} & \textbf{82.2} & \textbf{72.5} \\ \bottomrule
  \end{tabular}
  \label{hbrt_ablation}
\end{table}

\subsubsection{Comparison to Model of Other Online Tasks}

In Tab.\ref{compare_task}, as an important supplementary task for TAS, direct transferring TAS model to STAS cannot solve it, and the models of other tasks cannot perform well. This indicates that a unique model design is indeed needed for STAS task. Among the models shown in Tab.\ref{compare_task}, the image classification (IC) models completely lose the temporal information in the feature information. Although the Action Recognition (AR) models can use the temporal information in the video clip, they cannot detect the action boundary points, which makes the Acc score of AR slightly higher than IC, but the F1 scores is very low. Video Prediction (VP) models use historical information to predict future frames, and the loss of original feature information results in very poor experimental results. Although Online Action Detection (OAD) models can use the original feature information of the current frame while seeing historical feature information, it cannot guarantee the action segment integrity of full video, which makes it have a huge improvement over VP in terms of Acc, but it F1 scores is still very low.

\begin{table}[htbp]
\caption{Other online tasks comparison experiment in the split1 of GTEA.}
\centering
\scalebox{0.68}{
\begin{tabular}{@{}llllllll@{}}
\toprule
Publish   & Model                                                      & Task  & Acc  & Edit & \multicolumn{3}{l}{F1@\{0.1,0.25,0.5\}} \\ \midrule
CVPR 2016 & ResNet \cite{resnet}                      & IC    & 38.7 & 25.8 & 25.8        & 20.8        & 16.9        \\
CVPR 2018 & MobileNetV2 \cite{sandler2018mobilenetv2} & IC    & \textbf{40.9} & 32.2 & 30.5        & 23.6        & 14.2        \\
ICLR 2021 & ViT \cite{vit}                            & IC    & 28.7 & \textbf{33.6} & 21.9        & 17.3        & 8.3         \\
CVPR 2022 & Swinv2 \cite{liu2022swin}                 & IC    & 58.7 & 26.5 & \textbf{31.3}        & \textbf{27.6}        & \textbf{19.8}        \\
ICLR 2022 & MobileViT \cite{mehta2021mobilevit}       & IC    & 25.7 & 27.3 & 17.9        & 12.9        & 10.9        \\ \midrule
CVPR 2017 & I3D \cite{i3d}                            & AR    & 53.1 & 51.8 & 57.5        & 52.7        & 36.2        \\
CVPR 2018 & R(2+1)D \cite{r(2+1)d}                    & AR    & 24.4 & 36.6 & 30.0         & 24.3        & 13.4        \\
ICCV 2019 & TSM \cite{lin2019tsm}                     & AR    & 61.0 & \textbf{66.1} & 40.1        & 35.4        & 25.3        \\
ICML 2021 & TimeSformer \cite{timesformer}            & AR    & 36.4 & 31.7 & 29.3        & 24.3        & 18.6        \\
CVPR 2022 & Swin3D \cite{swin3d}                      & AR    & \textbf{63.4} & 60.2 & \textbf{65.1}        & \textbf{60.6}        & \textbf{48.2}        \\ \midrule
ICCV 2021 & OadTR \cite{wang2021oadtr}                & OAD   & 59.1 & 16.4 & 21.6        & 17.8        & 12.1        \\ \midrule
PAMI 2022 & PredRNNV2 \cite{wang2022predrnn}          & VP    & 22.7 & 27.9 & 18.8        & 17.7        & 13.5        \\ \midrule
BMVC 2021 & ASformer(full) \cite{yi2021asformer}      & TAS   & 75.9 & 84.3 & 86.2        & 83.4        & 75.6        \\ \midrule
BMVC 2021 & ASformer(streaming) \cite{yi2021asformer} & STAS & 70.7 & 68.7 & 76.9        & 70.7        & 57.1        \\
ours      & SVTAS-RL(MC)                              & STAS & \textbf{79.9} & \textbf{85.6} & \textbf{88.0}        & \textbf{86.5}        & \textbf{79.3}        \\ \bottomrule
\end{tabular}
}
\label{compare_task}
\end{table}

\subsubsection{Study on Memory Length}
We conducted experiments on the length of history information $m_j$, and the results are shown in Tab.\ref{memory_ablation}. SVTAS-RL achieved the best performance when the memory length is 512. As the $M$ continues to increase, SVTAS-RL continues to increase. When the memory length reaches 512, the performance is the best. When the $M$ reaches 1024, the overall performance no longer changes significantly. We speculate that this is because the time span of the dependency relationship between actions is mostly within 512.

\begin{table}[htbp]
\caption{The ablation study of $M$ in 50Salads.}
\centering
\scalebox{1.2}{
\begin{tabular}{@{}llllll@{}}
\toprule
$M$ & Acc & Edit & \multicolumn{3}{c}{F1@\{0.1,0.25,0.5\}} \\ \midrule
64         & 79.2 & 84.7 & 88.7 & 87.1 & 80.3 \\
128        & 77.8 & 85.5 & 89.7 & 87.7 & 80.4 \\
512        & 78.8 & \textbf{86.4} & \textbf{90.0} & \textbf{88.4} & 81.1 \\
1024       & \textbf{79.8} & 85.4 & 89.7 & \textbf{88.4} & \textbf{81.2} \\ \bottomrule
\end{tabular}
}
\label{memory_ablation}
\end{table}

\subsubsection{Study on Pre-training Parameters}
For the STAS task, we all used pre-trained weights as the parameters of VE, which can improve the performance of the model on STAS task. This is because the current datasets on TAS have a few samples and cannot provide sufficient training samples for VE. The ablation experiment results of pre-training parameters are shown in Tab.\ref{pre_traned}. In order to verify the original effectiveness of pre-training parameters, we conducted experiments on the Swin3D model. It can be seen that the experimental results using Kinetics-600 pre-training parameters are much better than those without using pre-training. In order to verify the effectiveness of pre-training parameters on our designed SVTAS model and select more effective pre-training parameters, we conducted experiments without using pre-training parameters, using SSv2 pre-training parameters and using Kinetics-600 pre-training parameters respectively. It can be seen that the experimental results without using pre-training parameters are far inferior to those using pre-training parameters, and the experimental results using Kinetics-600 pre-training parameters are better than those using SSv2 pre-training parameters. This is because SSv2 \cite{materzynska2020something} is a recognized dataset with strong temporal properties. VE with strong temporal modeling capabilities using SSv2 pre-training parameters cannot bring positive effects to STAS. Instead, VE trained by datasets such as Kinetics that can be recognized by image clustering can improve model performance. This proves that STAS is a clustering task.

\begin{table}[htbp]
\caption{Pre-trained experiment in 50Salads.}
\centering
\scalebox{.98}{
\begin{tabular}{@{}lllllll@{}}
\toprule
Pre-trained & Model & Acc  & Edit & \multicolumn{3}{c}{F1@\{0.1,0.25,0.5\}} \\ \midrule
 $\times$   & Swin3D& 38.9 & 21.4 & 23.7 & 18.4 & 12.6 \\
Kinetics-600& Swin3D& 84.7 & 59.6 & 69.6 & 67.9 & 61.0 \\
 $\times$   & SVTAS & 21.6 & 20.0 & 20.7 & 16.6 & 13.7 \\
  SSv2      & SVTAS & 85.1 & 75.1 & 82.7 & 80.5 & 73.9 \\
Kinetics-600& SVTAS & \textbf{86.7} & \textbf{78.4} & \textbf{85.3} & \textbf{83.7} & \textbf{77.7} \\ \bottomrule
\end{tabular}
}
\label{pre_traned}
\end{table}

\begin{figure}[hb]
  \centering
  % \fbox{\rule[-.5cm]{0cm}{4cm} \rule[-.5cm]{4cm}{0cm}}
  \includegraphics[width=\linewidth]{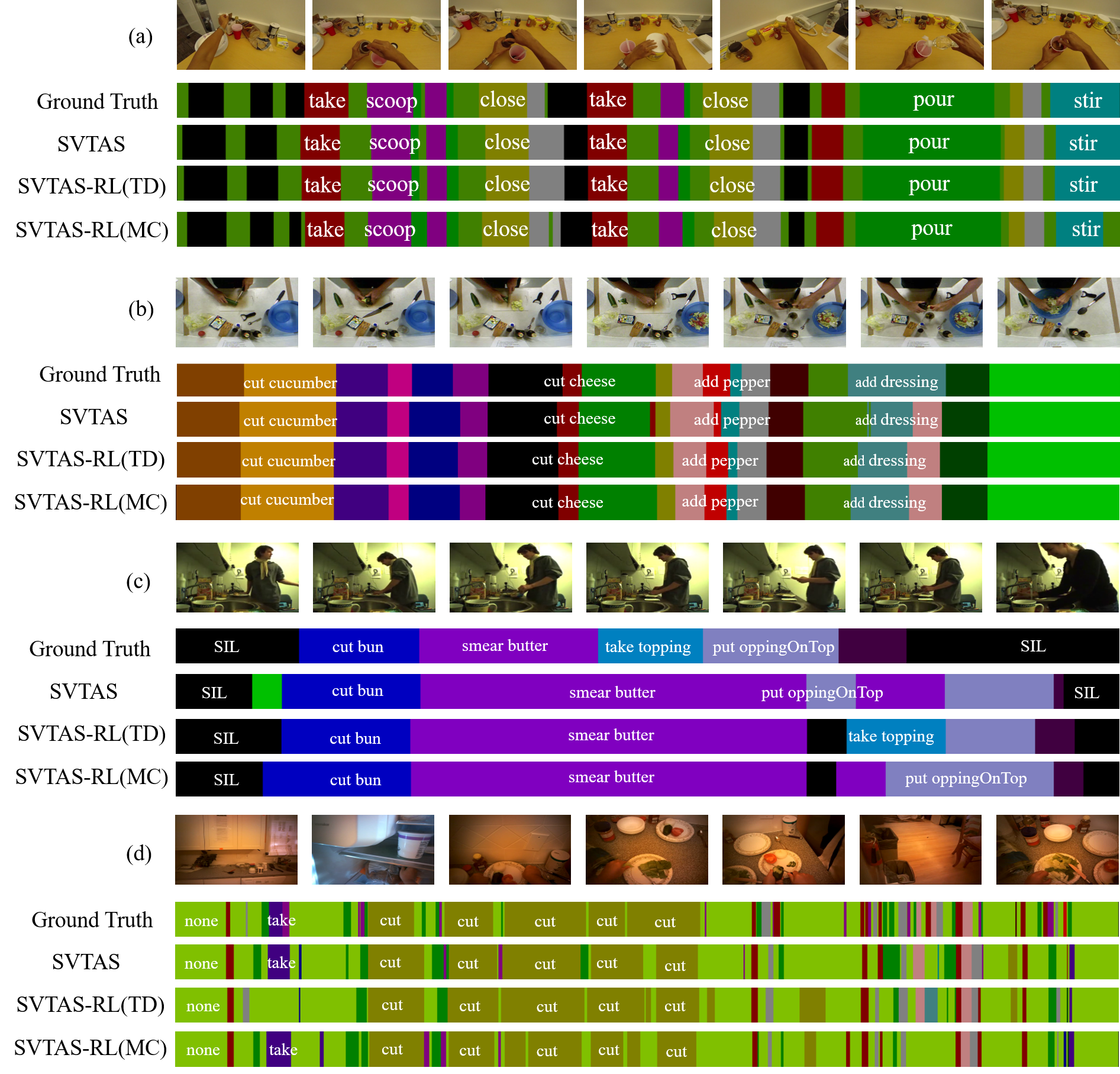}
  \caption{Quality results for datasets. (a) is from GTEA. (b) is from 50Salads. (c) is form Breakfast. (d) is from EGTEA.}
  \label{result_show}
\end{figure}

\subsection{Quality Results}
The quality results of our designed SVTAS, SVTAS-RL(TD), and SVTAS-RL(MC) on different datasets are shown in Fig.\ref{result_show}. It can be seen that the SVTAS model will have errors in action category recognition when performing segmenting on streaming videos. However, SVTAS-RL(TD) and SVTAS-RL(MC) using the RL training strategy can correct this error.

\section{Conclusion}
In the paper, we propose SVTAS-RL which eliminates modeling bias and alleviates optimization dilemma when TAS models migrate to STAS task. Specifically, we design SVTAS-RL based on clustering paradigm and introduce reinforcement learning training method by analyzing the modeling bias and optimization dilemma phenomenon. Extensive experiments have shown that STAS, as an important complementary task to TAS, has promising applications for processing long videos. Although our model still has a little bit latency when inferring, we believe that through our inspiration, the academic community will further explore to achieve real-time action segmentation.

\bibliographystyle{IEEEtran}
\bibliography{bare_jrnl_new_sample4.bib}

\end{document}